\documentclass[preprint,5p]{elsarticle}

\usepackage{amsmath,amssymb,amsfonts}
\usepackage{algorithmic}
\usepackage{graphicx}
\usepackage{textcomp}
\usepackage{rotating}   
\usepackage{multirow}
\usepackage{colortbl}
\usepackage{url}

\journal{arXiv}

\newcommand{\figref}[1]{Figure \ref{#1}}

\begin{document}

\begin{frontmatter}

\title{FEAFA: A Well-Annotated Dataset for Facial Expression Analysis and 3D Facial Animation}

\author[1]{Yanfu Yan}
\ead{yanyanfu16@mails.ucas.ac.cn}

\author[1]{Ke Lu}
\ead{luk@ucas.ac.cn}

\author[1]{Jian Xue\corref{cor}}
\ead{xuejian@ucas.ac.cn}

\author[1]{Pengcheng Gao}
\ead{gaopengcheng15@mails.ucas.ac.cn}

\author[2]{Jiayi Lyu}
\ead{lyujiayi\_cnu@163.com}

\cortext[cor]{Corresponding author}
\address[1]{School of Engineering Science, University of Chinese Academy of Sciences, Beijing 100049, China}
\address[2]{Information Engineering College, Capital Normal University, Beijing 100048, China}

\begin{abstract}
Facial expression analysis based on machine learning requires large number of well-annotated data to reflect different changes in facial motion. Publicly available datasets truly help to accelerate research in this area by providing a benchmark resource, but all of these datasets, to the best of our knowledge, are limited to rough annotations for action units, including only their absence, presence, or a five-level intensity according to the Facial Action Coding System. To meet the need for videos labeled in great detail, we present a well-annotated dataset named FEAFA for Facial Expression Analysis and 3D Facial Animation. One hundred and twenty-two participants, including children, young adults and elderly people, were recorded in real-world conditions. In addition, 99,356 frames were manually labeled using Expression Quantitative Tool developed by us to quantify 9 symmetrical FACS action units, 10 asymmetrical (unilateral) FACS action units, 2 symmetrical FACS action descriptors and 2 asymmetrical FACS action descriptors, and each action unit or action descriptor is well-annotated with a floating point number between 0 and 1. To provide a baseline for use in future research, a benchmark for the regression of action unit values based on Convolutional Neural Networks are presented. We also demonstrate the potential of our FEAFA dataset for 3D facial animation. Almost all state-of-the-art algorithms for facial animation are achieved based on 3D face reconstruction. We hence propose a novel method that drives virtual characters only based on action unit value regression of the 2D video frames of source actors. 
\end{abstract}


\end{frontmatter}

\section{Introduction}
\label{sec:introduction}
Humans communicate through the exchange of verbal and non-verbal messages, and facial expressions in fact play a key role in conveying non-verbal information between interactions. Therefore, facial expression analysis has become an increasingly important research area that can be used for many practical applications such as human-machine interaction \cite{Vinciarelli2009Social} and facial animation \cite{Cao2016Real}. Publicly available datasets are fundamental for accelerating research in facial expression analysis, because they not only require much effort to label massive numbers of video frames but also provide a benchmark through which researchers can compare their algorithm more objectively.

There are quite a few publicly available datasets for facial expression, such as the extended Cohn-Kanade (CK+) \cite{Lucey2010The}, DIFSA \cite{Mavadati2013DISFA}, MMI \cite{Bartlett2006Fully}, AM-FED \cite{Mcduff2013Affectiva}, SEMINA \cite{Mckeown2012The}, BP4D \cite{Zhang2013A}, and FERA \cite{Valstar2015FERA} datasets. All of these labeled datasets are based on the Facial Action Coding System (FACS) , which was initially developed in 1978 and was informed by earlier research. The FACS presents the most comprehensive catalogue of unique facial action units (AUs) and facial action descriptors (ADs) to describe all possible facial expressions. Each AU is anatomically based on facial actions that depend on a few facial muscles and may occur individually or in combination. In particular, ADs are not defined using specific facial muscles but can still represent some common facial motions. A recent version of FACS \cite{Ekman2002Facial}  assigns the letters A through E to represent AUs intensity variation, from barely detectable to maximum intensity, thus enabling the measurement and scoring of facial expression in an objective and quantitative way. However, the majority of the available FACS-coded datasets only label the presence or absence for AUs. Only few datasets provide the AU intensity at five levels. Even though the intensity information is accessible, these AU annotations with five-level can only suitable for detection tasks or rough intensity estimation tasks. There have been lots of state-of-the-art algorithms for AU detection \cite{Wu2017Deep,Li2017Action} and AU intensity estimation \cite{Walecki2016Copula,Walecki2017Deep} based on existing datasets. However, these existing datasets are not accurate enough to describe subtle intensity variation and enable an AU value regression task. 


\begin{figure}[t]
\centering
\includegraphics[width=0.95\columnwidth]{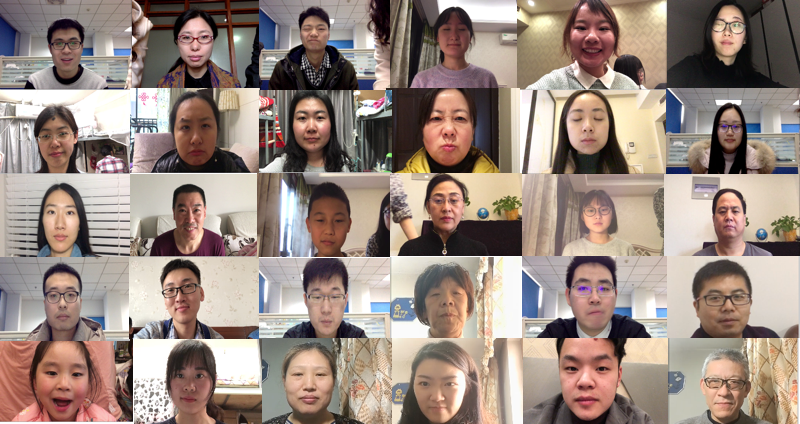}
\caption{Sample images from collected videos in the FEAFA dataset. These video sequences have various lightning, scale, position and pose. Participants express emotions including happiness, sadness, surprise, fear, contempt, anger and disgust.}
\label{fig:sample}
\end{figure}


The FACS has become a standard not only for the anatomically-based description of facial expression but also for muscle-based facial animation such as MPEG-4 facial animation parameters. Three-dimensional (3D) facial animation aims to capture the facial expression of source actors and then map the expression information to virtual characters. Impressive results in this area have been achieved, both based on RGB \cite{Cao2014Displaced,Cao2016Real,Thies2016Face2Face} and RGB-D \cite{Li2013Realtime} video data. However, almost all of these state-of-the-art techniques require a 3D shape reconstruction process, which is truly a time-consuming and tedious work. Because the most important information we want to obtain is facial expression parameters (to be more specific, AU values), we can in fact calculate them more conveniently from the 2D frames of a monocular video stream. Then we can use the AU values of the source actor and the corresponding avatar's blendshape for each AU to generate animations using linear combinations based on a blendshape model. The regression of AU values is thus necessary. Given that the facial appearance of some AU changes depending on the other AUs and some AUs co-occur more often than others, we provide a baseline system to achieve joint AU value regression. 

To meet the need of AU values regression, in this paper, we present a facial expression dataset named FEAFA with well-annotated AUs for the intensity with float numbers. The FEAFA dataset contains:

\begin{itemize}
\item \textbf{Facial videos:} 123 webcam videos of 122 participants recorded in real-world conditions.
\item \textbf{Labeled Frames:} 99,356 frames manually labeled using the Expression Quantification Tool to obtain the intensity as a floating point number between 0 and 1, for 9 symmetrical FACS AUs, 10 asymmetrical (unilateral) FACS AUs, 2 symmetrical FACS ADs and 2 asymmetrical FACS ADs. 
\item \textbf{Baseline System:} The Baseline performance of a joint AU value regression algorithm using deep Convolutional Neural Networks (CNNs) on this dataset. 
\item \textbf{New technique for 3D facial animation:} 3D facial animation of a virtual character that mimics the user's performance only using  the AU value regression of a 2D image rather than 3D shape reconstruction.
\end{itemize}

To facilitate future research on facial expression related algorithms, we have made the database and some demos available at \url{http://www.iiplab.net/feafa/}. Some sample images of the dataset are shown in \figref{fig:sample}. The rest of this paper is organized as follows: Section \ref{sec:existingdatasets} reviews some existing datasets related to our work; Section \ref{sec:thefeafa} describes the redefined AUs, data collection and annotation; Section \ref{sec:baseline} presents a baseline system to regress AU values; and Section \ref{sec:experiments} shows training, testing, and baseline performance of 24 regressed AU values on FEAFA. In Section \ref{sec:applications}, 
we demonstrate two major applications based on this dataset. And finally Section \ref{sec:conclusions} concludes the paper.


\section{Existing Datasets}
\label{sec:existingdatasets}
The Cohn-Kanade dataset \cite{Lucey2010The} plays a key role in the facial expression analysis. An extension of the Cohn-Kanade dataset (called CK+) contains 123 subjects and 593 posed and spontaneous sequences (10,708 frames), whose peak frames are AU coded with 30 AUs and given either intensity degree on a 7-point ordinal scale or an \textquotedblleft unspecified intensity.\textquotedblright~These videos are recorded under controlled light conditions and head motions.

The DISFA dataset \cite{Mavadati2013DISFA} contains 54 sequences of 27 young adults and participants. The participants generate facial expressions while watching emotive video clips. Each frame of the sequences (130,000 frames) was manually labeled for 12 AUs as presence, absence and intensity according the FACS. The naturalistic and spontaneous expressions of the facial behavior is imaged using a high resolution stereo-vision system under uniform illumination.

The MMI dataset \cite{Bartlett2006Fully} contains 25 subjects and 1,395 manually AU-coded sequences. The MMI labels consist of AU intensity for onset (start frame), apex (maximum intensity), and offset (end frame) but the dataset does not include intensity labels for each frame. MMI has many posed expressions and some spontaneous expressions, and both are recorded in a lab setting.

\begin{table*}[h]
\centering
\caption{AUs redefined for our dataset and the corresponding FACS AUs or FACS ADs. For AU43 (Eye Closure) in FACS, we regard Left Eye Close and Right Eye Close as two different AUs. In addition, AU2, AU4, AU5, AU20, AU30 in FACS are also subdivided in this way. For AU28 (Lip Suck) in the FACS, we subdivided it into Upper Lip Suck and Lower Lip Suck.}\label{tab:aulist}
\setlength{\tabcolsep}{1.8mm}
\small
\begin{tabular}{|c|l|l||c|l|l|}
\hline
 \textbf{AU} &  \textbf{Our Definition} & \textbf{FACS No. and Definition} &  \textbf{AU} &  \textbf{Our Definition} & \textbf{FACS No. and Definition} \\
\hline
\hline
1&Left Eye Close&AU43{ }{ }Eye Closure&13&Right Lip Corner Pull&AU12{ }{ }Lip Corner Puller\\
\hline
2&Right Eye Close&AU43{ }{ }Eye Closure&14&Left Lip Corner Stretch&AU20{ }{ }Lip stretcher\\
\hline
3&Left Lid Raise&AU 5{ }{ }{ }Upper lid raiser&15&Right Lip Corner Stretch&AU20{ }{ }Lip strecher\\
\hline
4&Right Lid Raise&AU 5{ }{ }{ }Upper lid raiser&16&Upper Lip Suck &AU28{ }{ }Lip Suck\\
\hline
5&Left Brow Lower&AU 4{ }{ }{ }Brow lowerer&17&Lower Lip Suck&AU28{ }{ }Lip Suck\\
 \hline
6&Right Brow Lower&AU 4{ }{ }{ }Brow lowerer&18&Jaw Thrust&AD29{ }{ }Jaw Thrust\\
 \hline
7&Left Brow Raise&AU 2{ }{ }{ }Outer brow raiser&19&Upper Lip Raise&AU10{ }{ }Upper Lip Raiser\\
 \hline
 8&Right Brow Raise&AU 2{ }{ }{ }Outer brow raiser&20&Lower Lip Depress&AU16{ }{ }Lower Lip Depressor\\
 \hline
  9&Jaw Drop&AU26{ }{ }Jaw Drop&21&Chin Raise&AU17{ }{ }Chin Raiser\\
  \hline
 10&Jaw Slide Left&AD30{ }{ }Jaw Sideways&22&Lip Pucker&AU18{ }{ }Lip Pucker\\
  \hline
 11&Jaw Slide Right&AD30{ }{ }Jaw Sideways&23&Cheeks Puff&AD34{ }{ }Puff\\
  \hline
  12&Left Lip Corner Pull&AU12{ }{ }Lip Corner Puller&24&Nose Wrinkle&AU 9{ }{ }{ }Nose wrinkler\\
\hline
\end{tabular}
\end{table*}

The AM-FED dataset \cite{Mcduff2013Affectiva} contains 242 AU related sequences (168,359 frames) of 242 subjects labeled frame by frame for the presence of 10 symmetrical AUs and four asymmetrical AUs. The AM-FED dataset records ecologically spontaneous facial responses online in real-world conditions. The SEMINA dataset \cite{Mckeown2012The} consists of 150 participants recorded in laboratory settings for a total of 959 conversations with individual SAL (Sensitive Artificial Listener) characters. Moreover, 180 images from eight sessions are FACS coded. The BP4D dataset \cite{Zhang2013A} contains 41 young adults generating spontaneous expression from stimulus tasks conducted in a lab environment. These expression sequences are labeled with the onset and offset of 27 AUs independently for each condition. The RU-FACS dataset \cite{Bartlett2006Automatic} contains video of the spontaneous behavior of 100 participants engaging in a 2.5 minute interview. The Bosphorus dataset \cite{Aly20083D} consists of 105 participants and the frames are labeled using FACS. The UNBC-McMaster Pain Archive \cite{Lucey2011Painful} provides naturalistic and spontaneous facial expressions with pain. This dataset is labeled for 10 AUs and the pain intensity is coded. The camera position, lighting, and resolution are controlled.

Most of the datasets reviewed above are coded for AU presence or absence and were collected in laboratory settings; only a few are coded with the five intensity levels introduced by the FACS and recorded in natural settings. To provide more accurate AU annotations, we hence create the FEAFA dataset based on videos recorded in real-world conditions.


\section{FEAFA Dataset}
\label{sec:thefeafa}
In this section, we redefine AUs based on the FACS, describe the collected video sequences, and show how the video frames are annotated in great detail.

\subsection{Action Unit Based on FACS}
We selected nine symmetrical FACS AUs, 10 asymmetrical (unilateral) FACS AUs, two symmetrical FACS ADs, and 2 asymmetrical FACS ADs to describe most expressions of the human face. To facilitate facial expression analysis based on our database, especially the blendshape process for 3D facial animation, we reorganized and renumbered the FACS AUs and ADs. Moreover, we refer to all of these facial actions as AUs for convenience. We also renamed the asymmetrical AUs, and, in particular, we also subdivided some AUs into upper and lower ones (e.g., Lip Suck). We hence obtained neutral expression and the 24 AUs to cover common facial expressions that vary among different individuals. The definition of each AU and its corresponding FACS AU or FACS AD are showed in Table \ref{tab:aulist}.

\subsection{Data Collection}
The facial behavior of 122 individuals (62 men and 60 women) was recorded by ordinarily monocular web cameras. We made Facial Expression Data Recording and Labeling Protocol (available at our website) for the participants to follow when their facial behaviors were recorded by the web cameras. This protocol, which will be available with the dataset at our website later, not only includes how to label the AUs in every image but also illustrates how to elicit the required expression. Besides, we trained every participant in order to avoid the problem of the mixture of children and adults. We required that facial area of the participants account for at least 100,000 pixels of the  frame and the orientation of the faces should be constrained, with at most a 30 degree deviation from a frontal view. In addition, these participants were all from Asia, and the facial expressions they presented were specified only through the already defined AUs regardless of whether the expression was a simple single expression or complex combination of  expressions. Furthermore, these videos were recorded under real-world conditions and thus exhibit non-uniform lighting and non-uniform frame rates. We obtained a total of 123 final sequences which last 7 - 112 seconds; each of the sequences consists of 4 - 29 different facial expressions.  From an AU value regression perspective, there are some challenges in our dataset such as the lighting of varied environment and differentiation of individuals' expressions, even if they present completely similar expressions. Additional details of our dataset are provided in Table \ref{tab:detail}.

\begin{table}[h]
\centering
\caption{Age distribution, glasses wearing, and facial hair occurrences in the video sequences in our FEAFA dataset.}\label{tab:detail}
\small
\begin{tabular}{|c|c|c|c|c|}
\hline
\multicolumn{3}{|c|}{\textbf{Age}}&\textbf{{ }Glasses{ }}&\textbf{Facial hair}\\
\hline
\hline
(0, 15]&(15, 60]&(60, 80]&Present&Present\\
\hline
4&110&8&40&38\\
\hline
 \end{tabular}
\end{table}

\begin{figure*}[htb]
\centering
\includegraphics[width=0.98\textwidth]{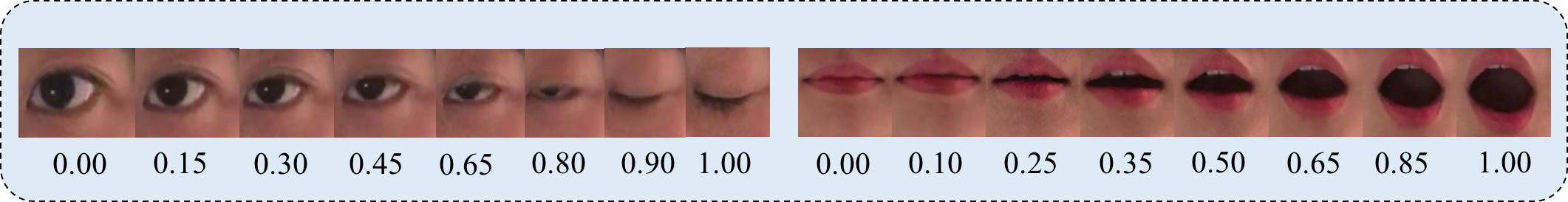}
\caption{Examples of AU2 (Right Eye Close) and AU9 (Jaw Drop) annotation for cropped facial images.}
\label{fig:annotation}
\end{figure*}


\subsection{AU Annotation}

Each of the posed sequences were independently labeled frame by frame. Every frame was coded by at least three highly trained coders chosen from 20 coders. These coders manually labeled each AU we already defined and they worked independently. The reliability was also tested through coding to ensure consistency. AU annotations were then subsequently labeled by another independent FACS-trained individual and discrepancies within the coding were reviewed.

We require labels that are more precise than those of FACS which only has five levels for each action units. Hence, we use floating point numbers from 0 to 1 and accurate to two decimal places to quantify each AU. A facial action state that is close to a neutral state is given a corresponding AU value close to 0; larger deviations of the facial action state from the neutral state are given a corresponding AU value that is closer to 1. This special annotation method is instrumental in identifying  AU values with expression coefficients. \figref{fig:annotation} gives examples of  AU2 (Right Eye Close) and AU9 (Jaw Drop) annotation for cropped facial images.

\begin{figure}[b]
\centering
\includegraphics[width=0.98\columnwidth]{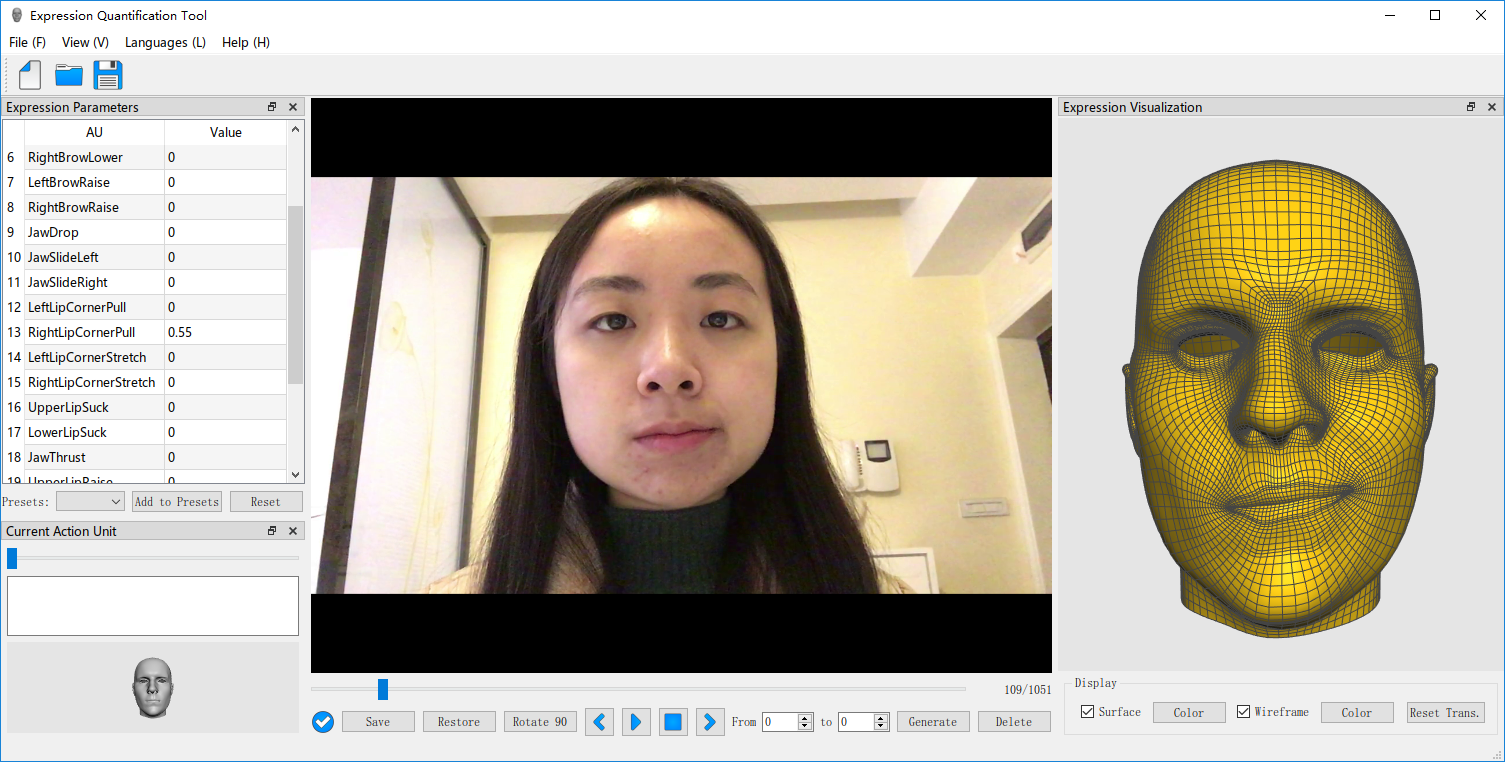}
\caption{Screenshot of the ExpreQuantTool used to label the video sequences on the FEAFA dataset.}
\label{fig:screenshot}
\end{figure}

To label AUs faster and more conveniently, we developed a software called Expression Quantification Tool (ExpreQuantTool), which is specifically designed for labeling data more efficiently. \figref{fig:screenshot} shows a screenshot of our software. On the right side of the software's GUI, the generated expression blendshape of a virtual character mimicking the expression of the subject of the frame in the Expression Visualization window is shown. 

\begin{figure}[t]
\centering
\includegraphics[width=0.98\columnwidth]{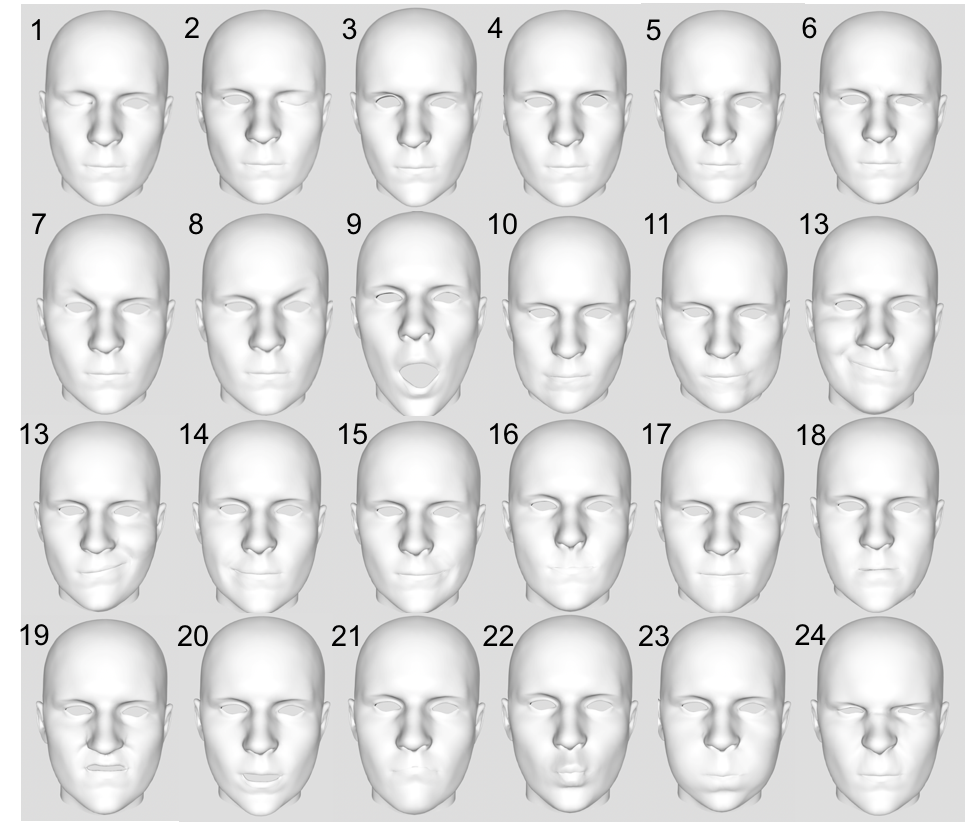}
\caption{Example of the corresponding expression blendshape for each AU.}
\label{fig:blendshapes}
\end{figure}

Similar to many facial animation techniques, we represent facial expressions in terms of expression blendshapes. To generate any possible expression of the source actors, we need a neutral face blendshape and 24 expression blendshapes according to 24 predefined AUs as our base poses. \figref{fig:blendshapes} shows an example of the expression blendshapes selected from FaceWarehouse \cite{Cao2014FaceWarehouse}.
Using these blendshapes $B = \{B_{0} , B_{1} , ..., B_{24}\}$, any facial expression of the subject of each recorded sequence can be easily replicated through linear combinations using the following equation.
\begin{equation}
\label{eq:blendshape}
	M=B_{0}+ \sum_{i=1}^{24}\beta_{i}(B_{i}-B_{0}),
\end{equation}
where $\beta =\{\beta_{1}, \beta_{2}, ..., \beta_{24}\}$ is a vector of expression coefficient and  $B_{0}$ denotes the neutral face.

When the coder labels the frame of certain expression of the video subjects displayed in the middle window, we can then obtain the AU values of the source actors, which actually are the expression coefficients of the virtual character, as well. Using the blendshapes and expression coefficients, we then generate the virtual character's specific expression model. This tool is of immense help to the coders to annotate images more precisely. During labeling, our ExpreQuantTool presents a corresponding 3D facial expression model once coders set AU values for a video frame. We can hence compare the expression on the 3D facial model with the expression on the source image directly so that the influence of the subjectivity on the AU labels accuracy can be minimized. Meanwhile, this tool can help to control differences of annotated AU values for one image between coders to a certain range. Besides, the AU values will be further validated to eliminate obviously wrong annotations and then we take the average of valid AU values among coders. After the coders finish labeling all the frames of one video sequence, they can play this video and the generated expression model based on the blendshapes will play automatically and is thus instrumental in double checking the labels of the frames.

This process yielded a total of 99,356 well-annotated frames. The distribution of each AU in four intervals is listed in Table \ref{tab:distribution}.

\begin{table}[]
\centering
\caption{Distribution of each AU on the FEAFA dataset. The range (0,1]  is evenly divided into four intervals to present our dataset in great detail.\label{tab:distribution}}
\small
\begin{tabular}{|c|c|c|c|c|}
\hline
 \textbf{AU} & \textbf{(0, 0.25] }&  \textbf{(0.25, 0.5]} &  \textbf{(0.5, 0.75]} & \textbf{(0.75, 1]} \\
\hline
1&13052&7105&2967&16199\\
\hline
2&12551&7163&2958&16263\\
\hline
3&1391&1190&960&2340\\
\hline
4&1384&1126&956&2396\\
\hline
5&3193&3044&2635&3359\\
\hline
6&2761&3248&2690&3423\\
\hline
7&2138&1816&888&3083\\
\hline
8&2141&1693&920&3210\\
\hline
9&7033&2581&2010&5720\\
\hline
10&921&889&552&1043\\
\hline
11&815&695&490&985\\
\hline
12&4726&4995&3023&2300\\
\hline
13&4800&4770&3082&2305\\
\hline
14&3492&3102&865&1186\\
\hline
15&3488&3084&895&1219\\
\hline
16&3237&2334&1649&3418\\
\hline
17&2860&1924&1285&3466\\
\hline
18&813&517&414&810\\
\hline
19&5321&3862&2319&3057\\
\hline
20&6942&5097&2512&3635\\
\hline
21&1706&1585&838&1730\\
\hline
22&2060&1917&1443&3275\\
\hline
23&657&669&435&2805\\
\hline
24&911&522&823&948\\
\hline

\end{tabular}
\end{table}


\section{Baseline System}
\label{sec:baseline}
\begin{figure*}[t]
\centering
\includegraphics[width=0.98\linewidth]{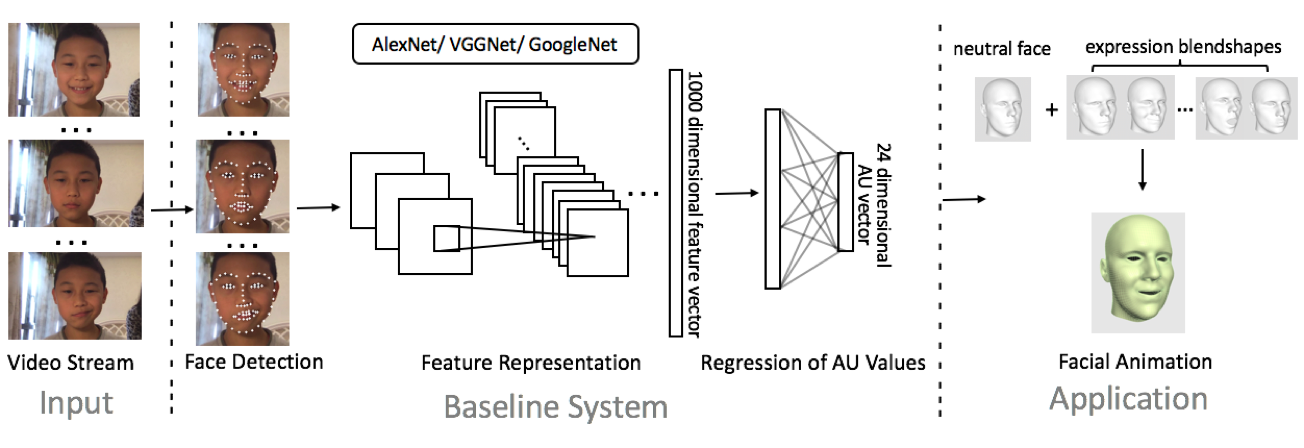}
\caption{Overview of baseline system and its application to 3D facial animation. The face image is obtained from the 2D video frames of an ordinary web camera and then tracked using the dlib library. The cropped image is passed through AlexNet, VGGNet and GoogleNet respectively to extract three facial features. These three features are used for regressing of AU values using a 3-layer neural network.}\label{fig:baseline}
\end{figure*}

Our baseline system consists of the complete pipeline of image pre-processing, feature representation, and AU value regression. We first use  the dlib detector based on histogram of oriented gradient (HoG) features to track the face. We then use the dlib C++ library to detect crucially fiducial points and crop the video frames. Each cropped face is then passed through deep CNN architectures to compute the features required for the regression of AU values. Finally, we use a 3-layer neural network based on the regression model to estimate the AUs. An overview of our baseline system is given in \figref{fig:baseline}. These modules are described in following subsections.

\subsection{Image Pre-processing}
HoG-based object detectors have become a classic computer vision method for detecting semi-rigid objects and have attracted attention in the face analysis field. Dlib's HoG detector \cite{King2009Dlib} can represent both appearance and shape information. It is relatively accurate at multiple resolutions and multiple pose angles and detecting faces in unconstrained scenarios. Hence, we use the HoG-based dlib detector to track the face.

After obtaining facial bounding box, we then use the dlib C++ library of \cite{Kazemi2014One}, which uses an ensemble of regression trees to locate the face's landmark positions from a sparse subset of intensity values in milliseconds with high quality predictions. It optimizes an appropriate loss function and performs feature selection in a data-driven manner, specifically using the gradient tree boosting method while minimizing the same loss function during training and test. The cascade regression can be summarized using following equation.
\begin{equation}
\label{eq:lib}
S^{(t+1)}=S^{(t)}+r_{t}(I, S^{(t)}),
\end{equation}
where $S$ denotes the facial shape. The shape estimate $S^{(t+1)}$ at the $(t + 1)^{th}$ stage can be written in terms of the shape estimate at the previous stage $S^{(t)}$ and shape regressor $r_{t}$, making its predictions based on features such as pixel intensity values. 

Once the facial landmarks have been detected, we calculate the distance between the center of two pupils, which is used to locate the face ROI. We then crop the face image and rescale it according to the input of the network for the feature representation.

\subsection{Feature Representation}
Various publicly available CNN architectures that have been trained on enormous datasets can be used as feature extractors. Once we obtain the cropped face image, we compute the facial features using AlexNet, VGGNet, and GoogleNet. We describe these CNNs in detail in the following subsections. For all three networks, we extract deep features from the last fully connected layer, which results in three feature vectors of length 1,000.

\subsubsection{AlexNet}
AlexNet \cite{Krizhevsky2012ImageNet} requires 227 $\times$ 227 RGB images as input, which are obtained in the image pre-processing step. AlexNet is a relatively shallow network, consisting five convolutional (Conv) layers and three fully-connected (FC) layers. It employs kernels with large receptive fields in layers close to the input and smaller kernels close to the output, incorporates rectified linear units(ReLU) instead of the hyperbolic tangent as the activation function and uses normalization layers and the overlapping pooling technique to aid generalization.

\subsubsection{VGGNet}
VGGNet \cite{Simonyan15} not only performs well in image classification tasks but also generates better visual representations. VGGNet is a large network that requires input size 224 $\times$ 224 $\times$ 3 pixels. Here, We use a 16-layer model often referred to as VGG-16 to represent facial features. VGG-16 includes 13 Conv layers, five max-pooling layers and three FC layers. It employs a much deeper network than AlexNet by stacking smaller kernels (3 $\times$ 3) instead of using a single layer of kernels with a large receptive field. The convolution stride is fixed to 1 pixel and the max-pooling layers are performed over a 2 $\times$ 2 pixel window with a stride of 2.

\subsubsection{GoogleNet}
GoogleNet \cite{7298594} employs a complex 22-layer deep network and uses inception modules to replace the mapping of convolutions of different sizes. The size of the receptive field is 224 $\times$ 224 taking RGB color channels with mean subtraction. Modules of the above type are stacked on each other, with occasional max-pooling layers with of stride 2 to halve the resolution of the grid. In addition, 1 $\times$ 1 convolutions compute reductions before expensive 3 $\times$ 3 and 5 $\times$ 5 convolutions. In addition, moving from FC layers to average pooling while retaining the use of dropout increases the top-1 accuracy.

\subsection{Regression of AU Values}
The values of facial AUs can be regressed independently and jointly using deep learning techniques, as done in many AU recognition methods . In contrast to the independent AU values regression, which ignores the relationship between AUs, the joint method can achieve better performance by adding AU relationships or dynamic dependencies.

\begin{figure}[t]
\centering
\includegraphics[width=0.9\linewidth]{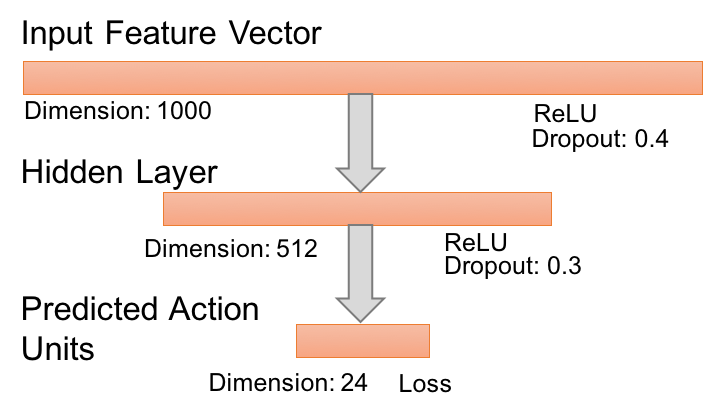}
\caption{Structure and parameters of the 3-layer neural network used for predicting the AU values.}
\label{fig:regression}
\end{figure}

We thus use a 3-layer neural network to jointly learn the AU values regression function. To balance real-time calculation with test accuracy, the number of layers was determined experimentally to be three. We use the back propagation algorithm with batch gradient descent to train the network weights and biases. Besides, unlike many AU detection and AU intensity estimation algorithms that evaluate the loss independently for each action unit, we regard them as a whole and evaluate all the loss from 24 AUs together. According to the Mean Square Error (MSE), which is commonly used to measure regression performance \cite{Rudovic2015Context}, we optimize the following loss function to learn the regression.
\begin{equation}
\label{eq:loss}
L_{train}=\frac{1}{N}\sum_{i=1}^{N}\Vert\textbf{x}_{i}-\hat{\textbf{x}}_{i}\Vert^{2},
\end{equation}
where $L_{train}$ is the average loss for all the training samples, $\hat{x}_{i}$ denotes the ground truth values of the AUs and $x_{i}$ denotes the predicted values of the AUs and is a vector of length 24. The first two FC layers are equipped with ReLU non-linearity. Because not all the dimensions are necessary for obtaining the information of the AUs, dimensionality reduction is needed, which is also beneficial to generalization. We thus add dropout \cite{Srivastava2014Dropout} after each FC layer so that the network itself can determine the best number of dimensions for the AU value regression rather than reducing the dimension manually. The dropout ratios are 0.4 and 0.3 for the input layer and the hidden layer, respectively. Over-fitting is thus reduced. The input layer accepts the 1000-dimensional arrays obtained from the facial feature representation task. After the ReLU and dropout operations, the output from the first layer is a hidden layer consisting of 512 hidden units. The following layer is an output layer containing 24 units that represent the regressed values of the 24 AUs and the loss between the ground truth labels and predicted labels is subsequently calculated. The 3-layer neural network and its parameters are showed in \figref{fig:regression}.

\begin{figure*}[htb]
\centering
\includegraphics[width=0.98\linewidth]{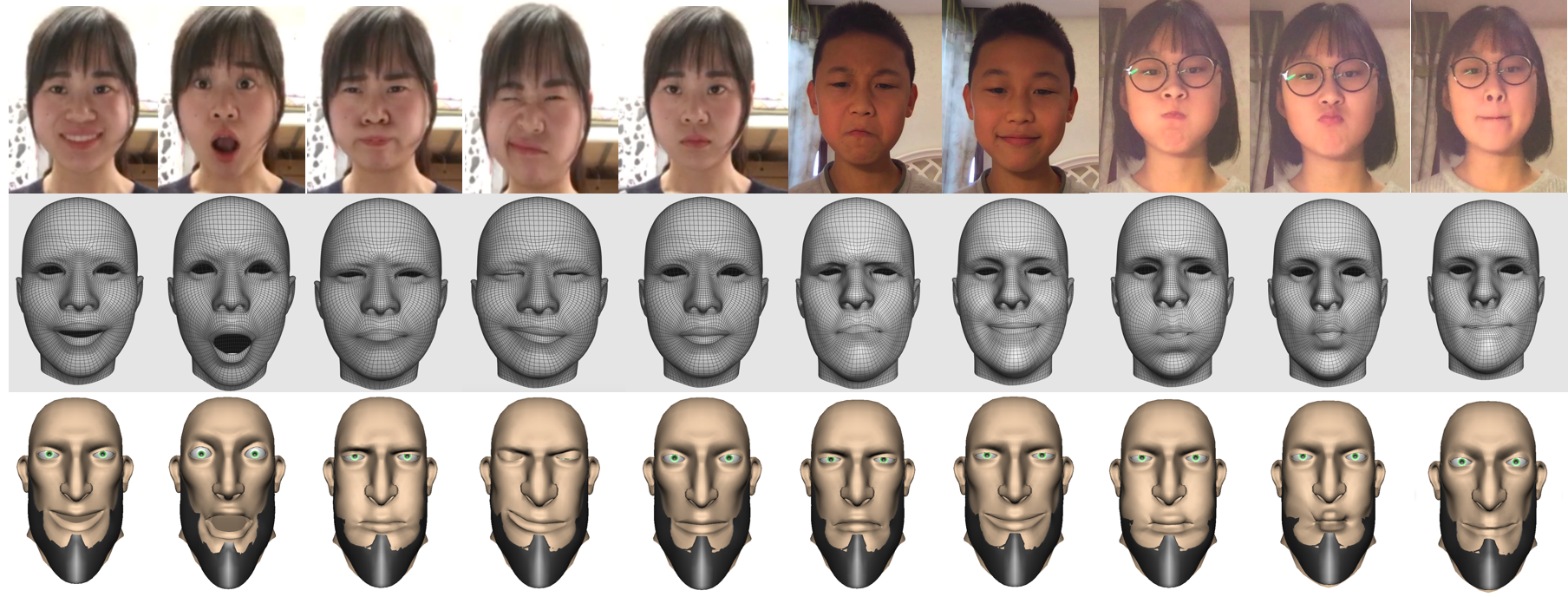}
\caption{Results for the facial animation of the simple virtual characters for ten frames.}
\label{fig:expression}
\end{figure*}


\section{Experiments}
\label{sec:experiments}
In this paper, we provide a baseline performance for the regression of the AU values conducted on our FEAFA database. Our method was implemented using the public Caffe framework \cite{Jia2014Caffe}. Training was performed on a NVIDIA Titan X Pascal GPU machine with 12GB of video memory. As described above, we used three feature representations from AlexNet, VGG-16, and GoogleNet and then used a 3-layer neural network to regress the values of the 24 AUs. 

We adopted a standard five-fold subject-exclusive cross-validation protocol \cite{Eidinger2014Age} to augment the numbers of training and testing data. Each AU in our dataset was distributed uniformly within these five folds. In the validation stage, the network parameters were selected by minimizing the loss function mentioned in \eqref{eq:loss}. During validation, the hyperparameters, such as learning rate, weight decay, and number of iterations, were optimized.

Network parameter (weight and bias) initialization strategies do have a significant effect on not only the outcome of the optimization procedure but also the ability of the network to generalize. Because the commonly used strategy, which initialize randomly from Gaussian or uniform distribution, only employ less information about the image, we believe that the pre-trained CNN architectures provide us with more information. Thus, the deep features extracted from the highest fully connected layers of AlexNet, VGG-16 and GoogleNet pre-trained for image classification were used for our AU value regression task. We directly used the bvlc\_alexnet model and bvlc\_googlenet model publicly provided in Caffe framework and downloaded the publicly available pre-trained network weights for the VGG\_ILSVRC\_16\_layers model provided by Visual Geometry Group. We then fine-tuned the network and obtained the final results of the values of AU regression. The regression results of using three CNN features respectively with or without pre-trained models are showed in Table \ref{tab:result}. The results show that pre-trained models give us a better result with lower loss than the original models.

Most importantly, the baseline performance indicates that the regression of AU values is feasible on this well-annotated and naturalistic dataset. However, there also remains room for improvement, such as how to regress AU values more accurately in low illumination conditions.

\begin{table}[]
\centering
\caption{Regression results pretained using three CNN features respectively on the FEAFA dataset for all 24 AUs. VGG-16 performs better than GoogleNet and AlexNet, whether the model is pre-trained or not.}\label{tab:result}
\small
\begin{tabular}{|c|c|c|}
\hline
\multicolumn{2}{|c|}{}&\textbf{Feature Extractors}\\
\multicolumn{2}{|c|}{}&ALexNet{ }{ }{ }VGG-16{ }{ }{ }GoogleNet\\
\hline
\multirow{2}{*}{\begin{sideways}MSE\end{sideways}}&not pre-trained&0.454{ }{ }{ }{ }{ }{ }{ }{ }{ }\textbf{0.398}{ }{ }{ }{ }{ }{ }{ }{ }{ }0.412\\
\cline{2-3}
&pre-trained&0.364{ }{ }{ }{ }{ }{ }{ }{ }{ }\textbf{0.289}{ }{ }{ }{ }{ }{ }{ }{ }{ }0.314\\
\hline
 \end{tabular}
\end{table}

\section{Applications}
\label{sec:applications}
The FEAFA dataset and its baseline system can be employed in various facial expression applications. In this section, we introduce two major applications: one is 3D facial animation from video sequences and the other is  AU detection and AU intensity estimation.

\subsection{3D Facial Animation from Video Sequences}
In this application, a virtual character is driven by the continuously changing facial performance of the source video sequence. We first use the algorithm described above to extract the AU values of the 2D video frames. The AU values of the source actor are later used as the blendshape coefficients of a digital avatar. In particular, we do not need to estimate the face identity of the participants in the video clip. Then, we can use an avatar's neutral face blendshape and the 24 corresponding expression blendshapes as basis poses. These blendshapes can be easily found in the FaceWarehouse \cite{Cao2014FaceWarehouse}, a database of 3D facial expression models containing the data of 150 individuals. Moreover, any virtual characters with 25 corresponding blendshapes can be used in this application. After obtaining the values of AUs from the 2D video frames and the corresponding blendshapes of the digital avatar, the animation of the digital avatar is generated by using a blendshape model through the linear combination given in \eqref{eq:genavatar}. In addition, we can also easily estimate the rigid transformation of head movements \cite{Lepetit2009EP}. Then the 3D facial mesh is represented as a linear combination of the basis blendshapes plus a rotation $R$ and translation $t$. It can be summarized by the following equation.
\begin{equation}
\label{eq:genavatar}
	F=R(B_{0}+\sum_{i=1}^{24}\beta_{i}(B_{i}-B_{0}))+t.
\end{equation}

This whole process is more convenient using AU value regression instead of 3D facial shape reconstruction. We also developed a software using features from AlexNet for facial animation. Our software runs at  over 18 fps on an macOS PC with an Intel (R) Core (TM) i5 CPU@3.1 GHz processor. In addition, because the AU value regression process takes AU relationships and dynamic dependencies into consideration, animation of the avatars is more verisimilar and naturalistic.  In \figref{fig:expression}, we present the results of the facial animation of simple virtual characters with combined AUs.

\subsection{AU Detection and AU Intensity Estimation}
There are quite a few state-of-the-art techniques for AU detection \cite{Wu2017Deep,Li2017Action} and AU intensity estimation \cite{Walecki2016Copula,Walecki2017Deep} based on deep learning. Binary AU detection clarifies whether facial motions are present or absent. AU intensity estimation recognizes not only whether the facial motions are present or not but also five increasing levels of facial expression intensity. Our FEAFA dataset is also suitable for these two tasks. For AU detection, we first use our baseline system to regress the AU values. And we regard each AU as presence only if its value is larger than 0.1. With regard to AU intensity estimation, after obtaining the regressed AU values, we divide the range (0,1] of AU values into (0, 0.2], (0.2, 0.4], (0.4, 0.6], (0.6, 0.8], and (0.8, 1] to represent the letters A through E, which describe an AUs intensity variation, from barely detectable to maximum intensity according to the FACS. Here, a certain AU is not present if its value is exact 0. 

\section{Conclusions and Future Work}
\label{sec:conclusions}
The majority of previous datasets for facial expression analysis are limited to rough AU annotations or were collected in lab settings. We present a facial expression dataset of FACS-labeled data that is exact in decimal. And we provide a parameter system with redefined AUs for facial expression quantization in order to describe facial expression in great detail. Our FEAFA dataset contains 122 participants from Asia, 123 videos recorded in real-world conditions, and 99,356 frames. Each frame is anatomically labeled by the presence of 9 symmetrical FACS AUs, 10 asymmetrical (unilateral) FACS AUs, 2 symmetrical FACS ADs and 2 asymmetrical FACS ADs. The intensity of each AU is annotated with a floating point number between 0 and 1. The FEAFA dataset thus provides continuous annotations of changes in facial expression. Through FEAFA, continuous AU values of any video frames could be regressed based on trained CNNs. These AU values can then be used to generate 3D facial animation easily and smoothly with great reality and perform expression recognition and classification faster and more precisely. Continuous AU values can quantify facial expression more accurately than discrete AU levels and enable more precise AU detection and AU intensity estimation. In the future, we hope the release of this dataset will encourage researchers to test new AU value regression algorithms that performs better than the benchmark we have already provided. For AU detection and AU intensity estimation, we expect that researchers can compare their state-of-the-art algorithms with each other using FEAFA. We also expect that many other applications other than 3D facial animation can benefit from our dataset, such as facial expression recognition and facial image manipulation.

\section*{Acknowledgements}
This work is supported by National Key R\&D Program of China under contract No. 2017YFB1002203, National Natural Science Foundation of China (NSFC, Grant No. 61671426, 61731022, 61871258, 61572077, 61471150), the Beijing Natural Science Foundation (Grant No. 4182071), the Instrument Developing Project of the Chinese Academy of Sciences (Grant No. YZ201670), and the Innovation Practice Training Program for College Students of Chinese Academy of Sciences.

\bibliographystyle{elsarticle-num}
\bibliography{refs}

\end{document}